\documentclass{article}
\usepackage{ijcai25}

\usepackage{times}
\usepackage{soul}
\usepackage{url}
\usepackage{subfigure}
\usepackage[hidelinks]{hyperref}
\usepackage[utf8]{inputenc}
\usepackage[small]{caption}
\usepackage{graphicx}
\usepackage{amsmath}
\usepackage{amsthm}
\usepackage{booktabs}
\usepackage{algorithm}
\usepackage{algorithmicx}
\usepackage{algpseudocode}
\usepackage{amssymb}
\usepackage[switch]{lineno}
\usepackage{pifont}
\usepackage{soul}
\usepackage{xcolor}

\definecolor{darkred}{HTML}{E63946}
\definecolor{lightblue}{HTML}{A8DADC}
\definecolor{darkblue}{HTML}{457B9D}

\usepackage{tikz} 
\usetikzlibrary{patterns}

\DeclareMathOperator{\argmin}{argmin} 

\title{Semi-Clairvoyant Scheduling of Speculative Decoding Requests to Minimize LLM Inference Latency}

\author{
 Ruixiao Li$^1$
\and
Fahao Chen$^2$\And
Peng Li\footnote{Corresponding author}$^1$\\
\affiliations
$^1$School of Cyber Science and Engineering, Xi'an Jiaotong University\\
$^2$School of Computer Science and Engineering, The University of Aizu\\
\emails
lrx4536257@stu.xjtu.edu.cn,
pengli@xjtu.edu.cn,
chenfh@ieee.org
}

\begin{document}

\maketitle

\begin{abstract}
    Speculative decoding accelerates Large Language Model (LLM) inference by employing a small speculative model (SSM) to generate multiple candidate tokens and verify them using the LLM in parallel. This technique has been widely integrated into LLM inference serving systems. However, inference requests typically exhibit uncertain execution time, which poses a significant challenge of efficiently scheduling requests in these systems. Existing work estimates execution time based solely on predicted output length, which could be inaccurate because execution time depends on both output length and token acceptance rate of verification by the LLM. In this paper, we propose a semi-clairvoyant request scheduling algorithm called Least-Attained/Perceived-Service for Speculative Decoding (LAPS-SD). Given a number of inference requests, LAPS-SD can effectively minimize average inference latency by adaptively scheduling requests according to their features during decoding. When the token acceptance rate is dynamic and execution time is difficult to estimate, LAPS-SD maintains multiple priority queues and allows request execution preemption across different queues. Once the token acceptance rate becomes stable, LAPS-SD can accurately estimate the execution time and schedule requests accordingly. Extensive experiments show that LAPS-SD reduces inference latency by approximately 39\% compared to state-of-the-art scheduling methods.
\end{abstract}

\section{Introduction}

Large Language Models (LLMs), such as the GPT series~\cite{brown2020language}, have demonstrated exceptional capabilities in various generative tasks~\cite{yao2024tree,zhuang2024toolqa}. LLM inference adopts an autoregressive decoding approach, which is inefficient because, for each token generated, it requires a full forward propagation of computing through the entire model. Recently, \textit{speculative decoding}~\cite{chen2023accelerating,leviathan2023fast} has been proposed as a promising approach to accelerate LLM inference. This technique leverages a small speculative model (SSM) alongside the primary LLM. The SSM first rapidly generates candidate tokens, which are then verified by the LLM. Since the SSM's compact size allows for high-speed generation of speculative tokens, and those tokens can be verified in parallel via a single forward pass of the LLM, speculative decoding achieves substantial inference speedups.

Due to the promising acceleration achieved by speculative decoding, existing work has integrated this technology into LLM inference serving systems to reduce inference latency~\cite{miao2024specinfer,li2024specpim,chen-infocom2025}. 
However, most existing work primarily focuses on developing advanced models to enhance the benefits of speculative decoding, while overlooking the critical challenge of inference request scheduling, which is essential to minimize the inference latency~\cite{patel2024splitwise,fu2024serverlessllm,llumix}. The main challenge of scheduling LLM requests lies in the unknown execution time of each request, as the number of its output tokens is uncertain. Some recent work~\cite{qiu2024efficient,zheng2024response} has proposed LLM output length prediction methods that can then be used to estimate execution time. For example, Qiu et al.~\cite{qiu2024efficient} employ a fine-tuned BERT-based model to predict the output length of inference requests, and Zheng et al.~\cite{zheng2024response} propose an instruction-tuned LLaMA model for the same purpose. With predicted output lengths, existing work adopts the Shortest-Job-First (SJF) algorithm to schedule requests to reduce the average inference latency, which is also called the average job completion time (JCT).

\begin{figure}[t]
    \centering
    \includegraphics[width=\linewidth]{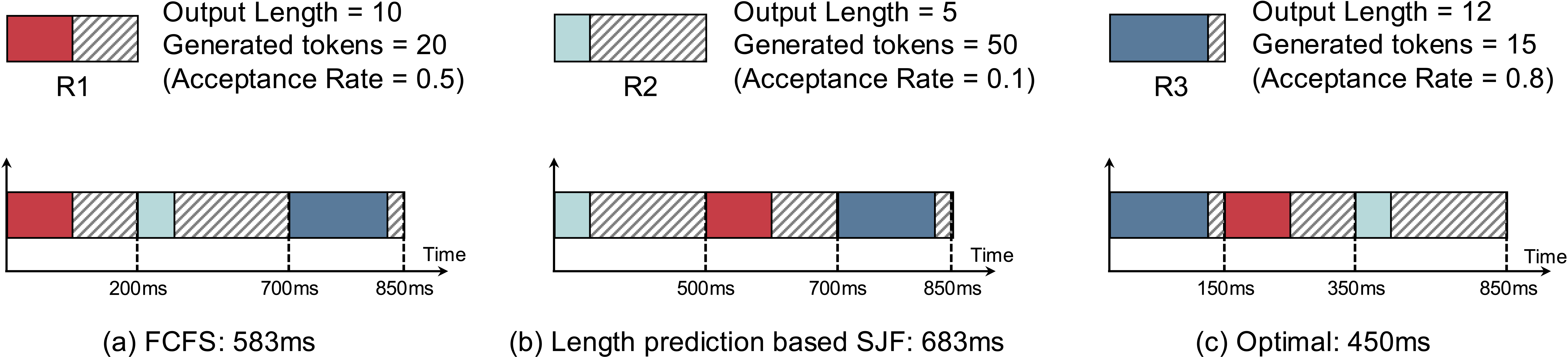}
    \caption{The illustration depicts different scheduling algorithms for speculative decoding requests. The generation context is represented by squares with colors (\textcolor{darkred}{$\blacksquare$} \textcolor{lightblue}{$\blacksquare$} \textcolor{darkblue}{$\blacksquare$}), while the speculative context is represented by squares with stripes.}
    \label{fig:motivation}
\end{figure}

However, in LLM serving systems using speculative decoding, relying solely on output lengths cannot accurately estimate inference execution time because it depends on both the output length and token acceptance rate of LLM verification, i.e., the proportion of tokens generated by the SSM that are accepted by the LLM during decoding. Specifically, some speculative tokens generated by the SSM could be rejected by the LLM but still contribute to the execution time. The total number of generated and verified tokens is generally larger than the output length. Consequently, using only output lengths for request scheduling could not effectively reduce the average inference latency.

\autoref{fig:motivation} provides a simple example to show how token acceptance rate affects request scheduling. In this example, three requests (R1, R2, and R3) have different output lengths and token acceptance rates. For example, R1 needs to generate 10 tokens as the final output, with an acceptance rate of 0.5. This implies that the SSM needs to generate 20 candidate tokens, which all need to be verified by the LLM. More candidate tokens mean that LLM needs to run more inference operations for token verification, which leads to a longer execution time.
We assume that three requests arrive in sequence at almost the same time, and each token requires an average of 10ms for verification by the LLM~\cite{miao2024specinfer}. For simplicity, the generation time of candidate tokens by the SSM is omitted due to its small size. In \autoref{fig:motivation}(a), we illustrate the scheduling results of the First-Come-First-Serve (FCFS) algorithm, which is commonly used by LLM inference serving systems~\cite{li2023alpaserve,kwon2023efficient}. Here, FCFS first schedules R1 and R2, with longer execution time, resulting in an average inference latency of 583ms. \autoref{fig:motivation}(b) shows the scheduling results of the existing SJF algorithm based on the predicted output lengths. R2 is scheduled first because it has the shortest output length of 5. However, R2 has the lowest acceptance rate of 0.1, taking the longest execution time to generate its 5 tokens. This leads to an average inference latency of 683ms. If we have information about both the request length and the acceptance rate, we can estimate the true execution time of speculative decoding requests, which leads to the optimal scheduling, as shown in \autoref{fig:motivation}(c).

In this paper, we propose a semi-clairvoyant request scheduling algorithm called Least-Attained/Perceptible-Service for Speculative Decoding (LAPS-SD), to minimize the average inference latency. LAPS-SD exploits a unique feature of speculative decoding that token acceptance rate is dynamic in the early stage of decoding and then becomes stable and predictable. Thus, LAPS-SD defines multiple execution priority queues and put requests in these queues according to their attained or perceptible inference service. In the early decoding stage when the acceptance rate is hard to be predictable, we assign priorities to requests according to their attained inference services. Execution preemption is allowed, but with negligible overhead because only a few tokens are generated in this stage. Later, as more tokens are generated and acceptance rate also becomes stable, LAPS-SD can accurately predict the total execution time and schedule requests by following the SJF principle. In such a way, LAPS-SD can well handle requests when their execution time is unknown, while reducing overhead of frequent preemption.


Our main contributions include:
\begin{itemize}
    \item We carefully examine the unique challenges in scheduling speculative decoding requests, and identify the weaknesses of existing work in minimizing inference latency. The obtained insights well motivate this paper.
    \item We propose a semi-clairvoyant scheduling algorithm, named Least-Attained/Perceived-Service for Speculative Decoding (LAPS-SD), which leverages both execution preemption and accurate execution time estimation to reduce inference latency.
    \item We evaluate LAPS-SD using three commonly used datasets: Chatbot Instruction Prompts~\cite{cip}, MBPP~\cite{austin2021program}, and MiniThinky~\cite{mini}. Extensive experiments demonstrate that LAPS-SD can reduce average inference latency by about 39\% compared to existing baselines.
\end{itemize}

The rest of this paper is organized as follows. \autoref{sec:background} provides the background, followed by the problem statement in \autoref{sec:problem}. The design of the scheduling algorithm is detailed in \autoref{sec:algorithm}, and the evaluation of the proposed algorithm is presented in \autoref{sec:evaluation}. \autoref{sec:related works} reviews related works, and \autoref{sec:conclusion} concludes the paper.

\section{Background}\label{sec:background}
\subsection{LLM Inference}\label{LLM Inference}
The inference process of Large Language Models (LLMs) is generally divided into two main stages. In the first stage, the entire prompt text is fed into the model to generate a KV cache and the first output logits. This process is efficient because it can process the entire prompt text in parallel. Let the prompt text be $x = \left[ x_1, x_2, \dots, x_n \right]$, where $x_i$ represents the $i-$th token. Upon receiving the prompt text, the model computes an initial hidden state $h_0$, and then generates the KV cache $KV = \{(k_1, v_1), (k_2, v_2), \dots, (k_m, v_m)\}$ based on this hidden state and the model parameters, where $k_i$ and $v_i$ represent the $i$-th key and value, respectively. Due to the parallel processing, the time complexity of this stage mainly depends on the number of layers in the model and the size of the hidden state, rather than the length of the prompt text.

The second stage is decoding and autoregressive generation, where the model generates tokens one by one. The model computes the next hidden state based on the current token and the previous hidden state. This process repeats until an termination token \texttt{<EOS>} is generated. Due to the sequential nature of the decoding process, generating each token requires streaming the entire model's weights through the computation units. Therefore, the arithmetic intensity, i.e., the ratio of floating-point operations (FLOPs) to memory bandwidth, of this stage is extremely low, especially when running with small batch sizes. This makes the decoding process typically the most expensive part of autoregressive generation~\cite{cai2024medusa,liu-etal-2024-speculative-decoding}.

\subsection{Speculative Decoding}\label{Speculative Decoding}

To accelerate the decoding process, speculative decoding uses an SSM to generate multiple candidate tokens, which are then used as prefixes along with the original input to be fed into a target LLM for parallel validation. For an input prefix sequence $X_{1:n} = \left[ x_1, x_2, \dots, x_n \right]$, the draft model autoregressively generates the subsequent $L$ tokens. Subsequently, the target LLM employs rejection sampling criteria to validate the generated candidate tokens.

The probability of a token $\hat{X}_{n+j}$ generated by the SSM can be represented as $p(\hat{X}_{n+j}| X_{1:n}, \hat{X}_{n+1:n+j-1})$. Then, the probability of the token $\hat{X}_{n+j}$ generated by the target LLM given a context $X_{1:n}, \hat{X}_{n+1:n+j-1}$ is $q(\hat{X}_{n+j}|X_{1:n}, \hat{X}_{n+1:n+j-1})$. So, the probability that the candidate token can be accepted is the minimum of the ratio of the probability distributions of the target LLM and the SSM for that token, but not exceeding 1, which can be formally expressed as: 
\begin{align}
    \min\left(1, \frac{q(\hat{X}_{n+j} | X_{1:n}, \hat{X}_{n+1:n+j-1})}{p(\hat{X}_{n+j} | X_{1:n}, \hat{X}_{n+1:n+j-1})}\right).
\end{align}

If the candidate token is rejected, a new token is resampled from the residual distribution. Therefore, the execution time of each speculative decoding request depends on both the output length and the token acceptance rate.

\subsection{LLM Inference Scheduling}\label{LLM Inference Scheduling}

With the widespread adoption of LLMs, e.g., GPT-3, LLaMA, in serving systems, efficient inference scheduling has become crucial for minimizing latency to ensure high service quality. However, scheduling remains challenging due to the uncertainty in the execution time of inference requests, and the complexity further increases when speculative decoding technology is integrated. Traditional First-Come-First-Serve (FCFS) scheduling can cause head-of-line blocking, leading to large inference latency.

Existing work estimates execution time by predicting request output lengths~\cite{qiu2024efficient,zheng2024response} to optimize inference scheduling performance. However, predicting output lengths does not accurately estimate the execution time of inference requests with speculative decoding. Specifically, the number of candidate tokens generated in speculative decoding is typically larger than the predicted request length, as some tokens may be rejected by the LLM. All candidate tokens must be verified by the LLM, which contributes to the total execution time. Therefore, existing methods that rely solely on predicted output lengths cannot accurately estimate execution time, leading to performance degradation in inference scheduling with speculative decoding.

In the absence of estimated execution time, some work adopts Least-Attained-Service (LAS)-based scheduling for inference requests~\cite{leviathan2023fast}. The key idea of LAS scheduling is to enable request preemption, ensuring that long-running requests do not block short ones. However, preemption introduces additional switching costs, primarily arising from the I/O overhead required to switch the KV pairs of different requests. As the inference progresses, these switching costs increase due to the growing size of KV pairs.

\section{Problem Statement}\label{sec:problem}
We consider an LLM inference serving system enabling speculative decoding. It receives inference requests, which are denoted by $\mathcal{N}$, and each request $i \in \mathcal{N}$ is associated with an arrival time $r_{i}$ and an execution time $T_{i}$. We assume that the batch size of this serving system is set to 1 for clear presentation. Note that our proposed algorithm can be easily extended to larger batch sizes. 

The newly arrived requests need to wait if the system is busy in serving other requests. We define a variable $x_{i}$ as the inference start time of request $i \in \mathcal{N}$ and the corresponding inference completion time is denoted by $C_{i}$. Thus, the inference latency can be calculated by $C_{i}-r_{i}$. If request execution cannot be preempted, the problem of minimizing the average inference latency can be formulated as follows: 
\begin{align} 
\text{min} \quad &\frac{1}{|\mathcal{N}|}\sum\limits_{i\in \mathcal{N}} (C_i-r_{i}), \quad \text{subject to:} \label{obj}\\
& x_{i} \geq r_{i}, \forall i \in \mathcal{N}; \label{cons_1} \\ 
& C_{i} \geq x_{i} + T_{i}, \forall i \in \mathcal{N}; \label{cons_2} \\
& |x_{i} - x_{j}|\geq T_{k}, \forall i,j \in \mathcal{N}, k\!=\! \argmin_{k\!=\!i,j}\{x_{k}\},\label{Cons_3}
\end{align} 
where the constraint (\ref{cons_1}) ensures that each request cannot start before its arrival time, and inference completion time is constrained by (\ref{cons_2}). We use the constraint (\ref{Cons_3}) to guarantee the non-preemption among inference requests.

\begin{figure}[t]
    \centering
    \includegraphics[width=0.9\linewidth,height=0.4\linewidth]{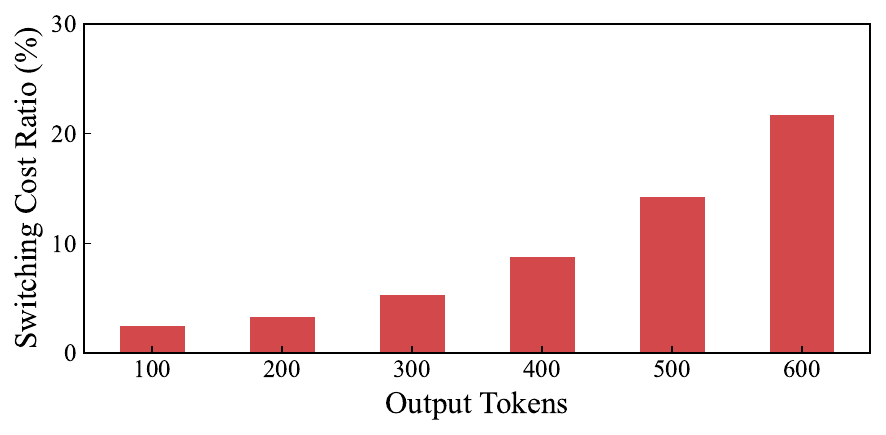}
    \caption{The ratio of switching costs to the inference time of requests with different output lengths.}
    \label{fig:switching_cost}
\end{figure}
\begin{figure}[t]
    \centering
    \includegraphics[width=\linewidth]{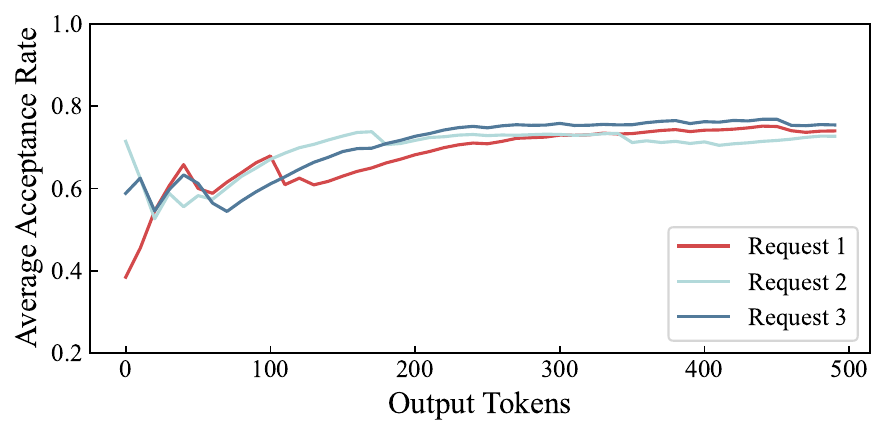}
    \caption{The average acceptance rate of three example requests over the speculative decoding process.}
    \label{fig:acceptance_rate}
\end{figure}

However, existing work fails to solve the above formulation because they lack prior knowledge of the execution time of each request, i.e., $T_{i}$, which depends on both output length and token acceptance rate. Different from output length that can be estimated before execution \cite{qiu2024efficient,zheng2024response}, token acceptance rate is hard to be predicted because its dynamics, as shown in \autoref{fig:acceptance_rate}, thus leading to unknown execution time. 
A common practice to handle unknown execution time is to use Least-Attained-Service (LAS) algorithm~\cite{rai2003analysis}, which gives the highest execution priority to the request received the least service time. LAS allows requests to be preempted, preventing long requests from blocking short ones. However, frequent preemption introduces significant overhead, primarily due to the I/O costs associated with switching the KV caches of different requests for LLM verification. The ratio of switching costs to total LLM inference time for requests with varying output lengths is shown in \autoref{fig:switching_cost}. We can see that switching requests with long output lengths during speculative decoding introduces significant overhead. For example, switching a request with an output length of 500 tokens adds a 14.21\% overhead to the total LLM inference time.

\section{Methodology}\label{sec:algorithm}
In this section, we present the proposed Least-Attained/Perceived-Service scheme for speculative decoding, called LAPS-SD, to address the weaknesses of existing works. An overview of LAPS-SD is given first, followed by elaboration of two key designs. 

\subsection{Overview}\label{Overview}
LAPS-SD is motivated by the observation that token acceptance rate is unstable in the early decoding stage and then stabilizes over execution. As shown in \autoref{fig:acceptance_rate}, we present the average token acceptance rates of three requests during the speculative decoding process. For every ten newly accepted tokens, calculate the average acceptance rate of all currently generated tokens. We observe that the acceptance rate is unstable during the early stages of decoding but stabilizes as decoding progresses. For example, request 2 has a relatively stable acceptance rate after generating 150 tokens.
Based on this observation, we are motivated to schedule speculative decoding requests as follows: in the early stage when token acceptance rates are difficult to predict, requests are scheduled by following the general LAS principle that allows execution preemption to avoid blocking. Once the acceptance rates of some requests become stable and predictable, they are scheduled by following the SJF principle without preemption to reduce execution switching cost.

\begin{algorithm}[t]
\caption{The Proposed LAPS-SD Scheduling Algorithm}
\label{alg:scheduling}
\begin{algorithmic}[1]
\Require Arrival speculative decoding requests;
\State Initialize priority queues;
\State Initialize requests' states as \textit{non-perceptible};
\State Put all requests in the queue with highest priority;
\State Schedule requests \texttt{InterQueueSchedule}();\label{PreempSchedule}
\Statex
\Procedure{\texttt{InterQueueSchedule}}{ } 
    \For{Non-empty queue with the highest queue}
        \State Schedule requests \texttt{IntraQueueSchedule}()
    \EndFor
\EndProcedure
\Statex
\Procedure{\texttt{IntraQueueSchedule}}{ }
    \If{Request becomes stable}
        \State Change request's state to \textit{perceptible};
        \State Predict the acceptance rate and request length;
        \State Estimate the execution time;
    \EndIf
    \State Schedule requests with semi-clairvoyant strategy;
\EndProcedure
\end{algorithmic}
\end{algorithm}

The pseudocodes of LAPS-SD are shown in \autoref{alg:scheduling}. It maintains multiple priority queues for request scheduling. For inter-queue scheduling (\autoref{Preemptive Scheduling}), requests in higher-priority queues are scheduled before those in lower-priority queues. 
Within each queue, requests are classified into two states: \textit{non-perceptible} and \textit{perceptible}. Non-perceptible requests lack predicted execution time information, and perceptible requests have such predictions because their acceptance rates become stable. Initially, all newly arrived requests are classified as non-perceptible ones and are placed in the highest-priority queue. Non-perceptible requests can be preempted if they have been executed for a certain period without completion, preventing longer requests from blocking shorter ones. At the same time, we monitor the acceptance rates of all requests to identify when they stabilize during execution. Once a request's acceptance rate becomes stable and predictable, it changes to the perceptible state and is moved to the corresponding queue accordingly, where they are scheduled with a semi-clairvoyant strategy to reduce the average inference latency (\autoref{perception scheduling}).

Non-perceptible requests can be well handled by LAPS-SD using different priority queues. Although execution preemption could incur switching cost of KV caches, the overhead is negligible because only a few tokens, whose KV cache is small, are generated for non-perceptible requests. As more tokens are generated, switching cost of KV caches becomes larger, and fortunately requests become perceptible and preemption is not allowed. Therefore, LAPS-SD can make full use of the strengths of LAS and SJF schemes while avoiding their weaknesses by exploiting the unique features of speculative decoding.

\subsection{Inter-Queue Scheduling}\label{Preemptive Scheduling}
\begin{figure}[t]
    \centering
    \includegraphics[width=\linewidth]{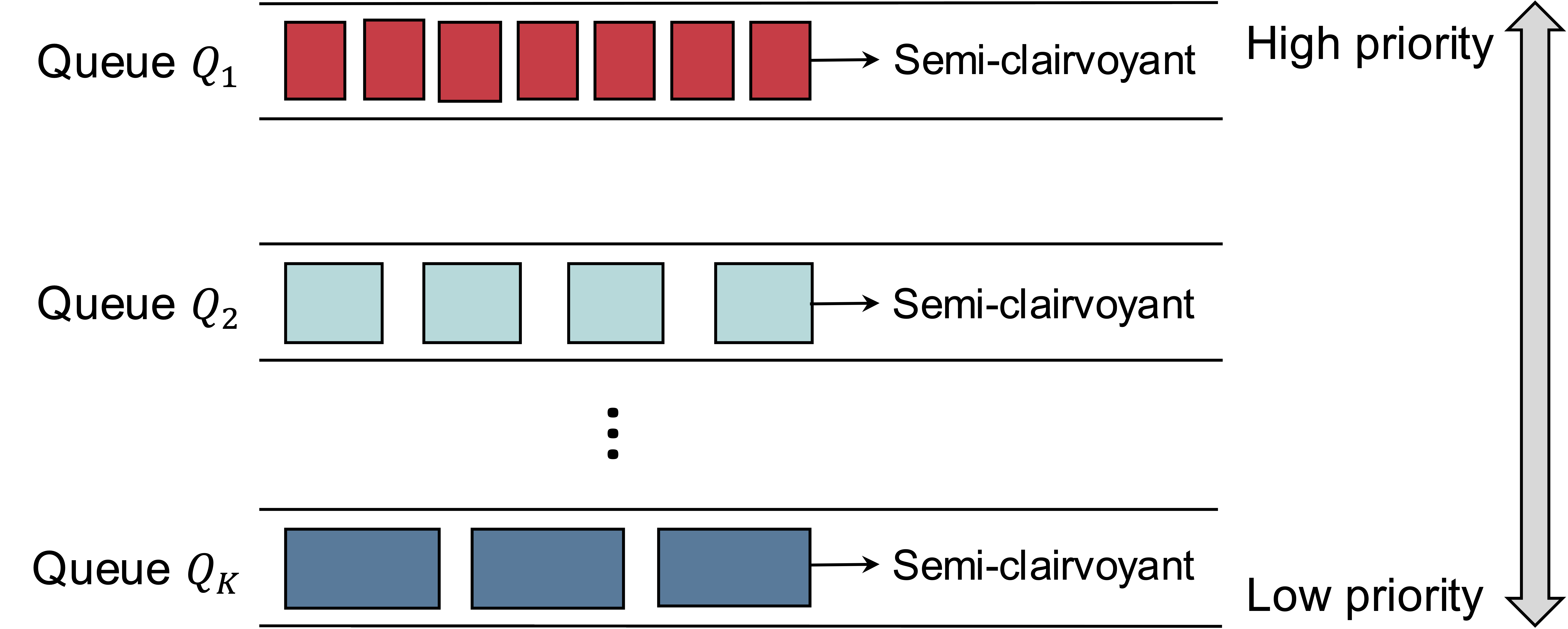}
    \caption{The queue structure in the proposed scheduling algorithm.}
    \label{fig:preemptive}
\end{figure}

As shown in \autoref{fig:preemptive}, we define $K$ priority queues, denoted as $\{Q_{1}, Q_{2}, \dots, Q_{K}\}$, where $Q_{1}$ has the highest priority and $Q_{K}$ has the lowest. Requests in a queue with higher priority are scheduled before those in lower-priority queues. Each queue $Q_{j}$ is associated with two thresholds, $S_{j}^{\text{down}}$ and $S_{j}^{\text{up}}$, which define the range of attained execution services that can be accommodated in the queue. Threshold of different queues exhibit an exponential relationship as $S_{j}^{\text{up}} = M^{j-1} \times S_{1}^{\text{up}}$. On the one hand, the exponentially increasing queue size enables us to handle requests with larger attained inference services using a smaller number of queues. On the other hand, queues for requests with large attained services will have a larger queue size. Typically, inference requests with large attained services correspond to longer output lengths, which involve large KV pairs. A larger queue size ensures that these requests are less likely to be preempted, thereby avoiding significant switching costs.
For a request $i$, its attained inference services $E_{i}$ is quantified by the current processing time, including both speculation and verification operations from the SSM and LLM.

The workflow of the preemptive scheduling is as follows. The priority of a speculative decoding request is determined by four lifecycle events:
\begin{itemize}
    \item \textbf{Arrival:} New requests are marked as non-perceptible state and are always placed in $Q_{1}$, the highest priority queue, since they have received no inference service.
    \item \textbf{Scheduling:} After each speculative decoding round, we calculate the accumulated execution time and demote the request to the appropriate queue according to the threshold $S_{j}^{down}$;
    \item \textbf{Stabilized:} Once the token acceptance rate of the request becomes stable and can be predicted, the request changes to the perceptible state;
    \item \textbf{Completion:} The request is moved out queues when it completes.
\end{itemize}

The overhead of maintaining these priority queues could be very low because it mainly involves statistical information (e.g., accumulated execution time) collection and state changes.

\subsection{Intra-Queue Scheduling}\label{perception scheduling}
Within each queue, we propose a multi-state scheduling strategy to schedule requests in different states. We first present how we predict the output length and acceptance rate. Then, we elaborate how the estimated execution time is derived based on these predictions. Finally, we introduce the semi-clairvoyant strategy.




\textbf{Output length and acceptance rate prediction.} To estimate the execution time accurately, we exploit both output length and acceptance rate. For the output length $L_{i}$ of request $i$, we adopt an existing method proposed by~\cite{zheng2024red}, which use a fine-tuned LLM model to predict the length of the generated response before scheduling. We predict the acceptance rate based on the observation that the rate gradually stabilizes over time. Specifically, we continuously monitor the acceptance rate for each request based on the total number of processed tokens and the number of accepted ones. When the maximum difference of the acceptance rate between $\gamma$ consecutive speculative decoding rounds is smaller than a given threshold $\delta$, the request $i$ is considered stable, and the average of acceptance rates in these $\gamma$ consecutive speculative decoding rounds is used as the predicted value $A_{i}$.

\textbf{Execution time estimation.} Based on the predicted output length $L_{i}$ and acceptance rate $A_{i}$, we estimate the execution time $\tilde{T}_{i}$ for request $i$ as follows. The speculative decoding of each request involves multiple rounds. In each round, the SSM autoregressively generates $n$ tokens, with the speculation time per token denoted by $T_{\text{LLM}}$. Then, the LLM verifies all generated speculative tokens in parallel, with a time cost of $T_{\text{SSM}}$. The estimated execution time $\tilde{T}_{i}$ for request $i$ is given by:
\begin{align}
    \tilde{T}_{i} = \Bigg\{\underbrace{ \frac{nL_{i}T_{\text{SSM}}}{n A_{i} + 1}}_{\text{Speculation Time}} + \underbrace{ \frac{L_{i}T_{\text{LLM}}}{n A_{i} + 1}}_{\text{Verification Time}}\Bigg\}
\end{align}
where $n A_{i} + 1$ indicates the estimated number of accepted tokens in a speculative decoding round (since the LLM always generates one additional token), and $\frac{L_{i}}{n A_{i} + 1}$ represents the number of rounds required to accept $L_{i}$ tokens. 

\textbf{Multi-state scheduling.} Since requests can arrive and stabilize at different times, they may be in different states even when placed in the same queue. In each queue, non-perceptible requests are scheduled using a FCFS strategy. Perceptible requests, which have an estimated execution time, are scheduled using an SJF strategy, as their request sizes are predictable. When handling requests with different states, we always prioritize scheduling perception requests. The rationale is as follows: non-perceptible requests have the potential to execute for a longer time, possibly exceeding the current queue threshold, while perceptible requests are less likely to exceed the queue's threshold. Therefore, prioritizing perceptible requests can benefit reducing the average inference latency.

\begin{figure*}[h]
\centering
        \subfigure[Chatbot Instruction Prompts.]{
		\begin{minipage}[b]{0.32\textwidth}
\includegraphics[width=1\textwidth]{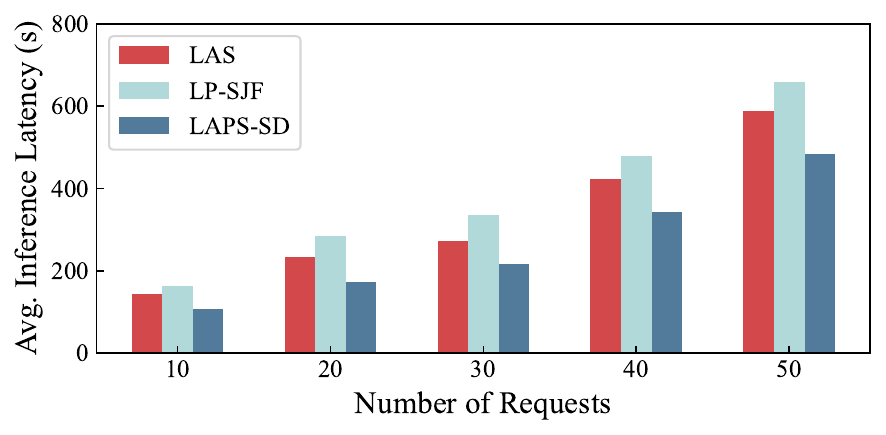}
        \end{minipage}
	\label{fig:overall_1}
	}
        \subfigure[MBPP.]{
            \begin{minipage}[b]{0.32\textwidth}
            \includegraphics[width=1\textwidth]{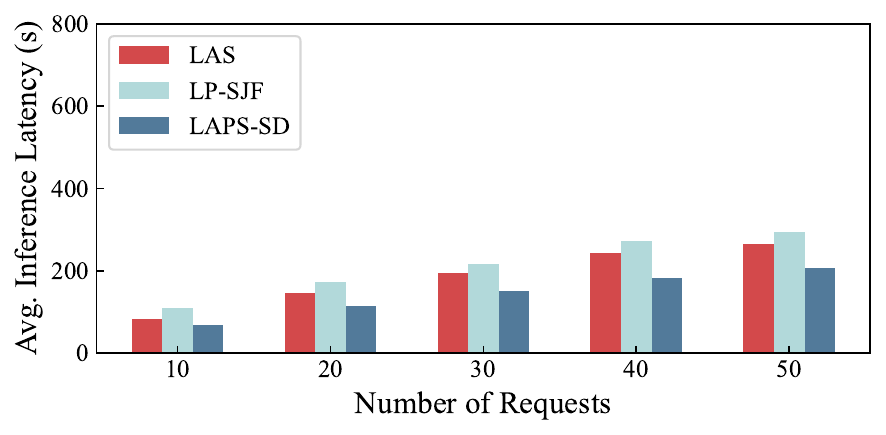}
            \end{minipage}
        \label{fig:overall_2}
        }
        \subfigure[MiniThinky.]{
            \begin{minipage}[b]{0.32\textwidth}
            \includegraphics[width=1\textwidth]{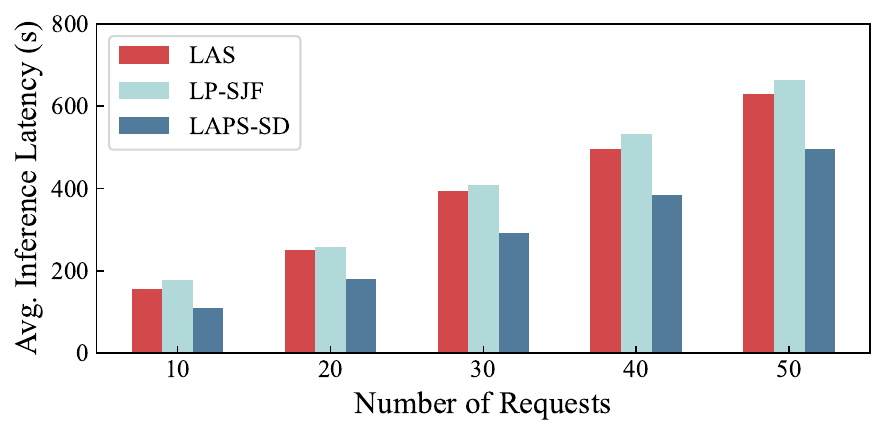}
            \end{minipage}
        \label{fig:overall_2}
        }
	\caption{The average inference latency with different scheduling algorithms.}
        \label{fig:overall}
\end{figure*}

\begin{figure*}[h]
\centering
        \subfigure[Chatbot Instruction Prompts.]{
		\begin{minipage}[b]{0.32\textwidth}
        
\includegraphics[width=1\textwidth]{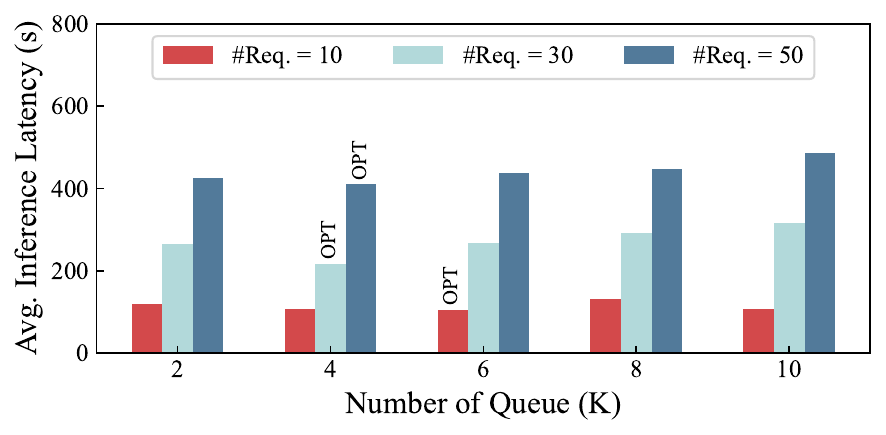}
        \end{minipage}
	\label{fig:k_1}
	}
        \subfigure[MBPP.]{
            \begin{minipage}[b]{0.32\textwidth}
            \includegraphics[width=1\textwidth]{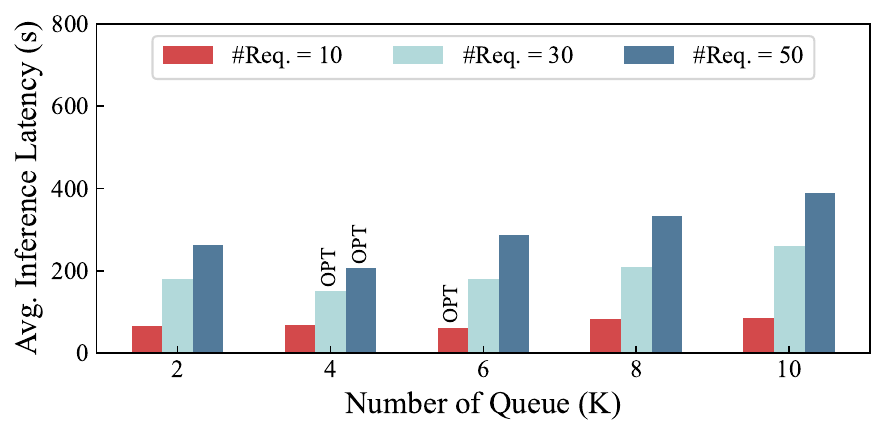}
            \end{minipage}
        \label{fig:k_2}
        }
        \subfigure[MiniThinky.]{
            \begin{minipage}[b]{0.32\textwidth}
            \includegraphics[width=1\textwidth]{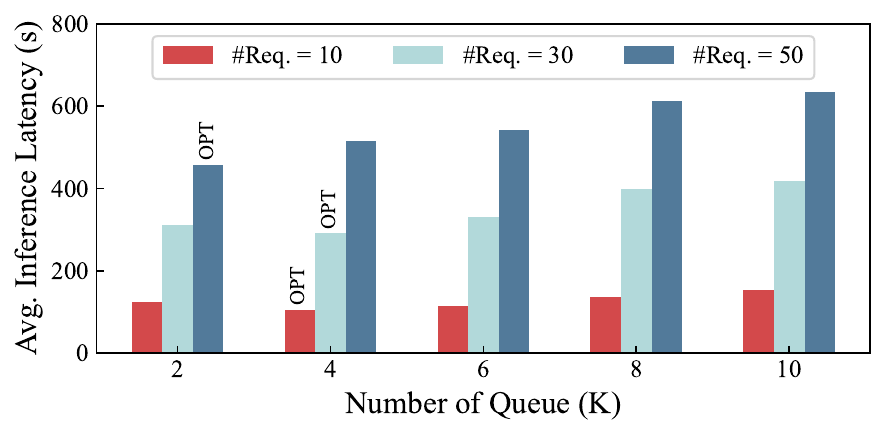}
            \end{minipage}
        \label{fig:k_3}
        }
	\caption{The impact of the number of priority queue.}
        \label{fig:dif_k}
\end{figure*}

\section{Evaluation}\label{sec:evaluation}

\subsection{Experiment Setup}\label{Experiment Setup}
\noindent\textbf{Environment.} We evaluate LAPS-SD on an NVIDIA L20 GPU with 48GB memory. The system runs Ubuntu 20.04.6 with Linux kernel version 5.15.0-91-generic, NVIDIA driver 550.120, CUDA 12.4, and cuDNN 8.6.0. The algorithm is implemented in Pytorch version 2.5.1.

\noindent\textbf{Workloads.} We evaluate our proposed scheduling algorithm using requests from three datasets: Chatbot Instruction Prompts~\cite{cip}, MBPP~\cite{austin2021program}, and MiniThinky~\cite{mini}, following the setup in \cite{miao2024specinfer}. We use LLaMA-68M~\cite{miao2024specinfer} as the SSM and the LLaMA-7B~\cite{touvron2023llama} as the LLM.

\noindent\textbf{Baselines.} We compare our scheduling algorithm against the following baselines: (1) Length Prediction-based Shortest Job First (LP-SJF)~\cite{qiu2024efficient}: this method uses the predicted output length to estimate the execution time and applies the SJF scheduling strategy; and (2) Least-Attained-Services (LAS): this method allows inference requests to be preempted based on their attained services, which has been adopted by~\cite{leviathan2023fast}. 

\begin{figure*}[h]
\centering
        \subfigure[Chatbot Instruction Prompts.]{
		\begin{minipage}[b]{0.32\textwidth}
\includegraphics[width=1\textwidth]{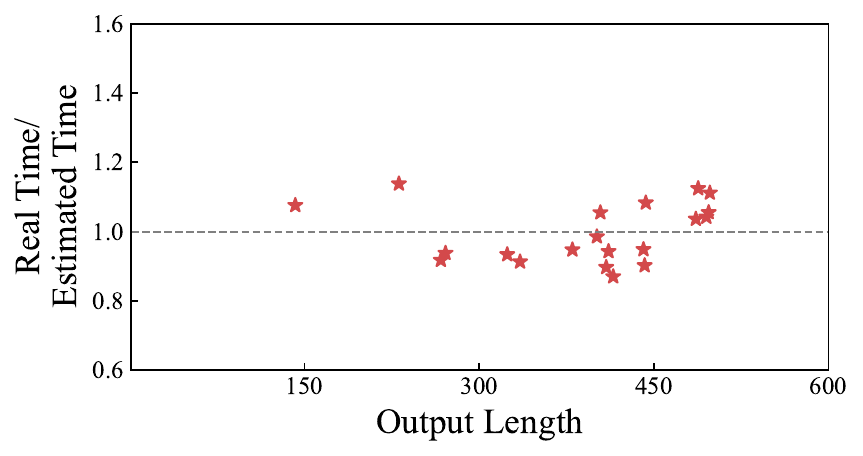}
        \end{minipage}
	\label{fig:acc_1}
	}
        \subfigure[MBPP.]{
            \begin{minipage}[b]{0.32\textwidth}
            \includegraphics[width=1\textwidth]{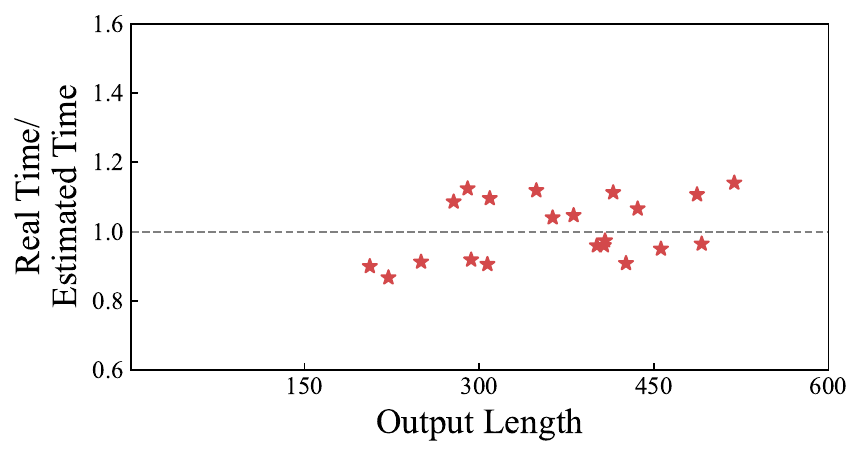}
            \end{minipage}
        \label{fig:acc_2}
        }
        \subfigure[MiniThinky.]{
            \begin{minipage}[b]{0.32\textwidth}
            \includegraphics[width=1\textwidth]{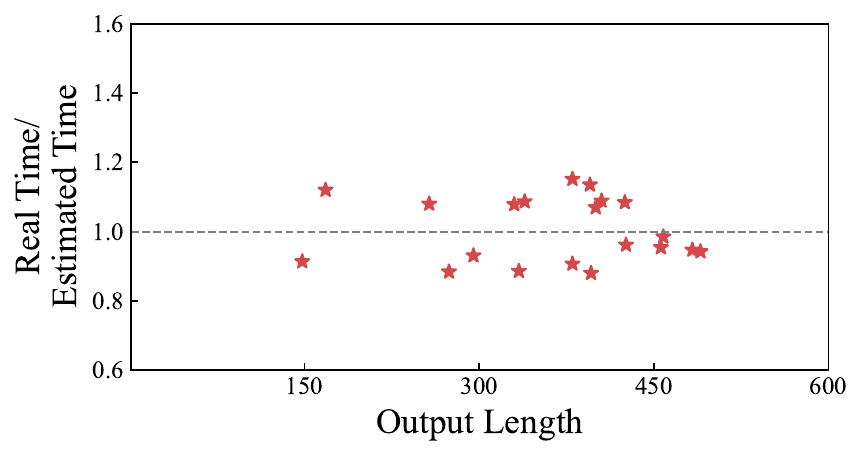}
            \end{minipage}
        \label{fig:acc_3}
        }
	\caption{The estimation accuracy of the execution time.}
        \label{fig:acc}
\end{figure*}

\subsection{Experiment Results}\label{Scheduling Performance}
We first evaluate the overall performance of our proposed scheduling algorithm by analyzing the average inference latency. As shown in \autoref{fig:overall}, the number of requests is varied from 10 to 50, and the inference latency of all requests increases as more requests are scheduled. Our proposed scheduling algorithm always outperforms others, reducing the average inference latency by about 39\%. Specifically, LP-SJF is approximately 1.47$\times$ slower than our proposed algorithm because relying solely on the predicted output length cannot accurately estimate execution time, resulting in higher inference latency. Compared to LAS, our scheduling algorithm reduces inference latency by about 31\%, as the estimated execution time effectively minimizes preemptions when requests are stable, thereby lowering the switching overhead. In addition, workloads from different datasets result in varying inference latencies. For example, the average inference latency for workloads from the MBPP dataset is significantly shorter than that of the Chatbot Instruction Prompts and MiniThinky datasets. This is because requests in the MBPP dataset have higher acceptance rates, leading to shorter processing latency for both the SSM and LLM.


We investigate the impact of the number of priority queues ($K$) on scheduling performance. Specifically, we change the number of queues from 2 to 10 and report the average inference latency of our proposed LAPS-SD algorithm in \autoref{fig:dif_k}. For requests from all datasets, we observe that the average inference latency initially decreases as the number of priority queues increases but eventually rises when the number of queues continues to grow. This behavior can be explained as follows: increasing the number of priority queues effectively prevents blocking by long requests and closely approximates the SJF strategy, thereby reducing the average inference latency. However, as the number of priority queues grows, preemption among requests becomes more frequent, introducing significant switching overhead. When the number of queues is small, the reduction in latency from mitigating long-request blocking outweighs the increased switching costs. Conversely, when the number of queues becomes large, the switching overhead dominates, resulting in increased inference latency.

Furthermore, we observe that the optimal number of priority queues varies with the number of inference requests. For example, as shown in \autoref{fig:k_1}, the optimal number of queues is 6 when there are 10 inference requests, but it decreases to 4 as the number of requests increases to 30. We analyze this as follows: with more inference requests, the possibility of preemption increases. As the number of queues grows, the associated switching costs also become more significant. Therefore, to mitigate the impact of frequent preemptions, maintaining a smaller number of queues becomes necessary for handling a larger number of inference requests effectively.

We also observe that inference requests from different datasets prefer different optimal numbers of priority queues. For example, with the same number of inference requests, i.e., 10, the optimal number of queues is 6 for requests from the Chatbot Instruction Prompts dataset, while it is 4 for requests from the MiniThinky dataset, as shown in \autoref{fig:k_2} and \autoref{fig:k_3}. This difference arises because the MiniThinky dataset has longer average input and output lengths, leading to higher switching costs for request preemption. As a result, a smaller number of queues is required to minimize the overhead for requests from the MiniThinky dataset.

We finally evaluate the effectiveness of the proposed method for estimating request execution time, by comparing the estimated execution time with the real execution time of inference requests under varying output lengths. The results are shown in \autoref{fig:acc}. Our proposed estimation method achieves an overall average error of 6.84\%, indicating the effectiveness of the proposed method for execution time estimation. In addition, the average estimation errors are 7.63\%, 11.21\%, and 8.51\% for the three datasets, respectively. The larger estimation error for requests on the MBPP dataset is due to the lower accuracy of the predicted acceptance rate compared to the other two datasets.

\section{Related Work}\label{sec:related works}

\subsection{Speculative Decoding}
Recently, speculative decoding~\cite{leviathan2023fast} has been proposed to accelerate LLM inference by adopting an SSM to generate multiple candidate tokens and then verify them with the LLM in parallel. Existing works strive to enhance the performance of speculative decoding. For example, 
Specinfer~\cite{miao2024specinfer} proposes a tree-based speculative inference and verification mechanism to reduce end-to-end latency. Glide with a Cape~\cite{duglide} reduces computational redundancy by leveraging an enhanced KV cache mechanism. N. Jha et al.~\cite{k} proposes incorporating speculative execution into the scheduling of control-flow-intensive designs, which can significantly improve performance. A staged speculative decoding algorithm~\cite{spector2023accelerating} accelerates LLM inference in small-batch scenarios. SpecDec++~\cite{huang2024specdec++} introduces an adaptive candidate length mechanism that dynamically adjusts candidate lengths to match the size of inference tasks and system load. Spectr~\cite{sun2024advance} optimizes the speculative decoding system using a verification framework based on optimal transport. Minions~\cite{wang2024minions} combines multiple inference tasks into a single batch, using grouped processing and pipeline mechanisms to improve system throughput. SmartSpec~\cite{liu2024optim} uses Goodput as a performance metric and implements priority-based scheduling. SpecExec~\cite{svirschevski2024specexec} proposes a massively parallel speculative decoding method designed specifically for resource-constrained consumer devices. Speculative Streaming~\cite{Nikhil2024streaming} implements a streaming framework that overlaps the token generation and verification processes. Although existing work has successfully improved the performance of speculative decoding, the scheduling problem for speculative decoding requests has been seldom studied, which motivates us to fill this gap in this paper.


\subsection{Scheduling for LLM Serving System}
In order to improve inference efficiency and performance, ExeGPT~\cite{oh2024exegpt} proposes a constraint-aware system that maximizes throughput while meeting latency requirements. PerLLM~\cite{yang2024perllm} introduces a personalized scheduling framework for edge-cloud collaboration. Another approach leverages LLMs to predict output lengths and group similar queries~\cite{zheng2024response}. Additionally, a learning-to-rank method for predicting relative output lengths~\cite{fu2024efficient} enables better approximation of shortest-job-first scheduling. FDIS~\cite{wu2023fast} decomposes inference tasks into smaller subtasks and processes them in parallel across multiple computing nodes to reduce latency and improve throughput. Adaptive Batch Budget~\cite{yecsil2024adaptive} presents an adaptive batch budget scheduling method to improve the efficiency of LLM inference by enhancing GPU utilization and throughput. Sarathi-Serve~\cite{agrawal2024taming} is an efficient LLM inference scheduler that improves serving throughput within desired latency SLOs by leveraging chunked-prefills to create stall-free schedules. INFERMAX~\cite{kim2024effect} analyzes that preemption mechanisms like LAS can reduce GPU costs. Existing scheduling methods primarily target traditional LLM inference requests, resulting in sub-optimal performance for speculative decoding requests. To address this, we leverage the perceptible characteristics of speculative decoding and propose a novel scheduling algorithm to minimize inference latency.


\section{Conclusion}\label{sec:conclusion}
In this paper, we present LAPS-SD, a semi-clairvoyant scheduling algorithm for LLM inference with speculative decoding. LAPS-SD combines execution preemption and execution time estimation to reduce inference latency. Specifically, LAPS-SD initially maintains multiple priority queues, allowing requests to be preempted when their execution times are difficult to predict, thereby preventing blocking issues caused by long requests. Once the execution times of requests become predictable, LAPS-SD estimates them accurately by predicting both output length and acceptance rate. Extensive experiments demonstrate that LAPS-SD reduces the average inference latency by approximately 39\% compared to baseline methods.

\section*{Acknowledgments}
This work was supported by the National Natural Science Foundation of China (No. 62471383).

\bibliographystyle{named}
\bibliography{ijcai25}

\begin{thebibliography}{}

\bibitem[\protect\citeauthoryear{Agrawal \bgroup \em et al.\egroup }{2024}]{agrawal2024taming}
Amey Agrawal, Nitin Kedia, Ashish Panwar, Jayashree Mohan, Nipun Kwatra, Bhargav Gulavani, Alexey Tumanov, and Ramachandran Ramjee.
\newblock Taming throughput-latency tradeoff in llm inference with sarathi-serve.
\newblock In {\em 18th USENIX Symposium on Operating Systems Design and Implementation (OSDI 24)}, pages 117--134, 2024.

\bibitem[\protect\citeauthoryear{{Alessandro Palla}}{2023}]{cip}
{Alessandro Palla}.
\newblock Chatbot instruction prompts.
\newblock \url{https://huggingface.co/datasets/alespalla/chatbot_instruction_prompts}, 2023.

\bibitem[\protect\citeauthoryear{Austin \bgroup \em et al.\egroup }{2021}]{austin2021program}
Jacob Austin, Augustus Odena, Maxwell Nye, Maarten Bosma, Henryk Michalewski, David Dohan, Ellen Jiang, Carrie Cai, Michael Terry, Quoc Le, et~al.
\newblock Program synthesis with large language models.
\newblock {\em arXiv preprint arXiv:2108.07732}, 2021.

\bibitem[\protect\citeauthoryear{Bhendawade \bgroup \em et al.\egroup }{2024}]{Nikhil2024streaming}
Nikhil Bhendawade, Irina Belousova, Qichen Fu, Henry Mason, Mohammad Rastegari, and Mahyar Najibi.
\newblock Speculative streaming: Fast llm inference without auxiliary models.
\newblock 2024.

\bibitem[\protect\citeauthoryear{Brown \bgroup \em et al.\egroup }{2020}]{brown2020language}
Tom Brown, Benjamin Mann, Nick Ryder, Melanie Subbiah, Jared~D Kaplan, Prafulla Dhariwal, Arvind Neelakantan, Pranav Shyam, Girish Sastry, Amanda Askell, et~al.
\newblock Language models are few-shot learners.
\newblock {\em Advances in neural information processing systems}, 33:1877--1901, 2020.

\bibitem[\protect\citeauthoryear{Cai \bgroup \em et al.\egroup }{2024}]{cai2024medusa}
Tianle Cai, Yuhong Li, Zhengyang Geng, Hongwu Peng, Jason~D. Lee, Deming Chen, and Tri Dao.
\newblock Medusa: Simple {LLM} inference acceleration framework with multiple decoding heads.
\newblock In {\em Forty-first International Conference on Machine Learning}, 2024.

\bibitem[\protect\citeauthoryear{Chen \bgroup \em et al.\egroup }{2023}]{chen2023accelerating}
Charlie Chen, Sebastian Borgeaud, Geoffrey Irving, Jean-Baptiste Lespiau, Laurent Sifre, and John Jumper.
\newblock Accelerating large language model decoding with speculative sampling.
\newblock {\em arXiv preprint arXiv:2302.01318}, 2023.

\bibitem[\protect\citeauthoryear{Chen \bgroup \em et al.\egroup }{2025}]{chen-infocom2025}
Fahao Chen, Peng Li, Tom~H. Luan, Zhou Su, and Jing Deng.
\newblock Spin: Accelerating large language model inference with heterogeneous speculative models.
\newblock In {\em Proceedings of IEEE International Conference on Computer Communications (INFOCOM)}, London, United Kingdom, May 2025.

\bibitem[\protect\citeauthoryear{Du \bgroup \em et al.\egroup }{2024}]{duglide}
Cunxiao Du, Jing Jiang, Xu~Yuanchen, Jiawei Wu, Sicheng Yu, Yongqi Li, Shenggui Li, Kai Xu, Liqiang Nie, Zhaopeng Tu, et~al.
\newblock Glide with a cape: A low-hassle method to accelerate speculative decoding.
\newblock In {\em Forty-first International Conference on Machine Learning}, 2024.

\bibitem[\protect\citeauthoryear{Fu \bgroup \em et al.\egroup }{2024a}]{fu2024serverlessllm}
Yao Fu, Leyang Xue, Yeqi Huang, Andrei-Octavian Brabete, Dmitrii Ustiugov, Yuvraj Patel, and Luo Mai.
\newblock Serverlessllm: Low-latency serverless inference for large language models.
\newblock In {\em 18th USENIX Symposium on Operating Systems Design and Implementation}, pages 135--153, 2024.

\bibitem[\protect\citeauthoryear{Fu \bgroup \em et al.\egroup }{2024b}]{fu2024efficient}
Yichao Fu, Siqi Zhu, Runlong Su, Aurick Qiao, Ion Stoica, and Hao Zhang.
\newblock Efficient llm scheduling by learning to rank.
\newblock In {\em The Thirty-eighth Annual Conference on Neural Information Processing Systems}, 2024.

\bibitem[\protect\citeauthoryear{Huang \bgroup \em et al.\egroup }{2024}]{huang2024specdec++}
Kaixuan Huang, Xudong Guo, and Mengdi Wang.
\newblock Specdec++: Boosting speculative decoding via adaptive candidate lengths.
\newblock {\em arXiv preprint arXiv:2405.19715}, 2024.

\bibitem[\protect\citeauthoryear{Kim \bgroup \em et al.\egroup }{2024}]{kim2024effect}
Kyoungmin Kim, Kijae Hong, Caglar Gulcehre, and Anastasia Ailamaki.
\newblock The effect of scheduling and preemption on the efficiency of llm inference serving.
\newblock {\em arXiv preprint arXiv:2411.07447}, 2024.

\bibitem[\protect\citeauthoryear{Kwon \bgroup \em et al.\egroup }{2023}]{kwon2023efficient}
Woosuk Kwon, Zhuohan Li, Siyuan Zhuang, Ying Sheng, Lianmin Zheng, Cody~Hao Yu, Joseph Gonzalez, Hao Zhang, and Ion Stoica.
\newblock Efficient memory management for large language model serving with pagedattention.
\newblock In {\em Proceedings of the 29th Symposium on Operating Systems Principles}, pages 611--626, 2023.

\bibitem[\protect\citeauthoryear{Lakshminarayana \bgroup \em et al.\egroup }{2000}]{k}
Ganesh Lakshminarayana, Anand Raghunathan, and Niraj~K Jha.
\newblock Incorporating speculative execution into scheduling of control-flow-intensive designs.
\newblock {\em IEEE Transactions on Computer-Aided Design of Integrated Circuits and Systems}, 19(3):308--324, 2000.

\bibitem[\protect\citeauthoryear{Leviathan \bgroup \em et al.\egroup }{2023}]{leviathan2023fast}
Yaniv Leviathan, Matan Kalman, and Yossi Matias.
\newblock Fast inference from transformers via speculative decoding.
\newblock In {\em International Conference on Machine Learning}, pages 19274--19286. PMLR, 2023.

\bibitem[\protect\citeauthoryear{Li \bgroup \em et al.\egroup }{2023}]{li2023alpaserve}
Zhuohan Li, Lianmin Zheng, Yinmin Zhong, Vincent Liu, Ying Sheng, Xin Jin, Yanping Huang, Zhifeng Chen, Hao Zhang, Joseph~E Gonzalez, et~al.
\newblock Alpaserve: Statistical multiplexing with model parallelism for deep learning serving.
\newblock In {\em 17th USENIX Symposium on Operating Systems Design and Implementation (OSDI 23)}, pages 663--679, 2023.

\bibitem[\protect\citeauthoryear{Li \bgroup \em et al.\egroup }{2024}]{li2024specpim}
Cong Li, Zhe Zhou, Size Zheng, Jiaxi Zhang, Yun Liang, and Guangyu Sun.
\newblock Specpim: Accelerating speculative inference on pim-enabled system via architecture-dataflow co-exploration.
\newblock In {\em Proceedings of the 29th ACM International Conference on Architectural Support for Programming Languages and Operating Systems, Volume 3}, pages 950--965, 2024.

\bibitem[\protect\citeauthoryear{Liu \bgroup \em et al.\egroup }{2024a}]{liu-etal-2024-speculative-decoding}
Jiahao Liu, Qifan Wang, Jingang Wang, and Xunliang Cai.
\newblock Speculative decoding via early-exiting for faster {LLM} inference with {T}hompson sampling control mechanism.
\newblock In {\em Findings of the Association for Computational Linguistics}, pages 3027--3043, 2024.

\bibitem[\protect\citeauthoryear{Liu \bgroup \em et al.\egroup }{2024b}]{liu2024optim}
X.~Liu, C.~Daniel, L.~Hu, W.~Kwon, Z.~Li, X.~Mo, A.~Cheung, Z.~Deng, I.~Stoica, and H.~Zhang.
\newblock Optimizing speculative decoding for serving large language models using goodput.
\newblock {\em arXiv preprint arXiv:2406.14066}, 2024.

\bibitem[\protect\citeauthoryear{Miao \bgroup \em et al.\egroup }{2024}]{miao2024specinfer}
Xupeng Miao, Gabriele Oliaro, Zhihao Zhang, Xinhao Cheng, Zeyu Wang, Zhengxin Zhang, Rae Ying~Yee Wong, Alan Zhu, Lijie Yang, Xiaoxiang Shi, et~al.
\newblock Specinfer: Accelerating large language model serving with tree-based speculative inference and verification.
\newblock In {\em Proceedings of the 29th ACM International Conference on Architectural Support for Programming Languages and Operating Systems, Volume 3}, pages 932--949, 2024.

\bibitem[\protect\citeauthoryear{Oh \bgroup \em et al.\egroup }{2024}]{oh2024exegpt}
Hyungjun Oh, Kihong Kim, Jaemin Kim, Sungkyun Kim, Junyeol Lee, Du-seong Chang, and Jiwon Seo.
\newblock Exegpt: Constraint-aware resource scheduling for llm inference.
\newblock In {\em Proceedings of the 29th ACM International Conference on Architectural Support for Programming Languages and Operating Systems, Volume 2}, pages 369--384, 2024.

\bibitem[\protect\citeauthoryear{Patel \bgroup \em et al.\egroup }{2024}]{patel2024splitwise}
Pratyush Patel, Esha Choukse, Chaojie Zhang, Aashaka Shah, {\'I}{\~n}igo Goiri, Saeed Maleki, and Ricardo Bianchini.
\newblock Splitwise: Efficient generative llm inference using phase splitting.
\newblock In {\em 2024 ACM/IEEE 51st Annual International Symposium on Computer Architecture (ISCA)}, pages 118--132. IEEE, 2024.

\bibitem[\protect\citeauthoryear{Qiu \bgroup \em et al.\egroup }{2024}]{qiu2024efficient}
Haoran Qiu, Weichao Mao, Archit Patke, Shengkun Cui, Saurabh Jha, Chen Wang, Hubertus Franke, Zbigniew~T Kalbarczyk, Tamer Basar, and Ravishankar~K Iyer.
\newblock Efficient interactive llm serving with proxy model-based sequence length prediction.
\newblock In {\em International Conference on Architectural Support for Programming Languages and Operating Systems}, 2024.

\bibitem[\protect\citeauthoryear{Rai \bgroup \em et al.\egroup }{2003}]{rai2003analysis}
Idris~A Rai, Guillaume Urvoy-Keller, and Ernst~W Biersack.
\newblock Analysis of las scheduling for job size distributions with high variance.
\newblock In {\em Proceedings of the 2003 ACM SIGMETRICS international conference on Measurement and modeling of computer systems}, pages 218--228, 2003.

\bibitem[\protect\citeauthoryear{Spector and Re}{2023}]{spector2023accelerating}
Benjamin~Frederick Spector and Christopher Re.
\newblock Accelerating {LLM} inference with staged speculative decoding.
\newblock In {\em Workshop on Efficient Systems for Foundation Models of ICML 2023}, 2023.

\bibitem[\protect\citeauthoryear{Sun \bgroup \em et al.\egroup }{2024a}]{llumix}
Biao Sun, Ziming Huang, Hanyu Zhao, Wencong Xiao, Xinyi Zhang, Yong Li, and Wei Lin.
\newblock Llumnix: Dynamic scheduling for large language model serving.
\newblock In {\em 18th USENIX Symposium on Operating Systems Design and Implementation (OSDI 24)}, pages 173--191, 2024.

\bibitem[\protect\citeauthoryear{Sun \bgroup \em et al.\egroup }{2024b}]{sun2024advance}
Ziteng Sun, Ananda~Theertha Suresh, Jae~Hun Ro, Ahmad Beirami, Himanshu Jain, and Felix Yu.
\newblock Spectr: Fast speculative decoding via optimal transport.
\newblock {\em Advances in Neural Information Processing Systems}, 36, 2024.

\bibitem[\protect\citeauthoryear{Svirschevski \bgroup \em et al.\egroup }{2024}]{svirschevski2024specexec}
Ruslan Svirschevski, Avner May, Zhuoming Chen, Beidi Chen, Zhihao Jia, and Max Ryabinin.
\newblock Specexec: Massively parallel speculative decoding for interactive {LLM} inference on consumer devices.
\newblock In {\em The Thirty-eighth Annual Conference on Neural Information Processing Systems}, 2024.

\bibitem[\protect\citeauthoryear{Touvron \bgroup \em et al.\egroup }{2023}]{touvron2023llama}
Hugo Touvron, Thibaut Lavril, Gautier Izacard, Xavier Martinet, Marie-Anne Lachaux, Timoth{\'e}e Lacroix, Baptiste Rozi{\`e}re, Naman Goyal, Eric Hambro, Faisal Azhar, et~al.
\newblock Llama: Open and efficient foundation language models.
\newblock {\em arXiv preprint arXiv:2302.13971}, 2023.

\bibitem[\protect\citeauthoryear{Wang \bgroup \em et al.\egroup }{2024}]{wang2024minions}
Siqi Wang, Hailong Yang, Xuezhu Wang, Tongxuan Liu, Pengbo Wang, Xuning Liang, Kejie Ma, Tianyu Feng, Xin You, Yongjun Bao, et~al.
\newblock Minions: Accelerating large language model inference with adaptive and collective speculative decoding.
\newblock {\em arXiv preprint arXiv:2402.15678}, 2024.

\bibitem[\protect\citeauthoryear{Wu \bgroup \em et al.\egroup }{2023}]{wu2023fast}
Bingyang Wu, Yinmin Zhong, Zili Zhang, Shengyu Liu, Fangyue Liu, Yuanhang Sun, Gang Huang, Xuanzhe Liu, and Xin Jin.
\newblock Fast distributed inference serving for large language models.
\newblock {\em arXiv preprint arXiv:2305.05920}, 2023.

\bibitem[\protect\citeauthoryear{{Xuan Son NGUYEN}}{2024}]{mini}
{Xuan Son NGUYEN}.
\newblock Minithinky dataset.
\newblock \url{https://huggingface.co/datasets/ngxson/MiniThinky-dataset}, 2024.

\bibitem[\protect\citeauthoryear{Yang \bgroup \em et al.\egroup }{2024}]{yang2024perllm}
Zheming Yang, Yuanhao Yang, Chang Zhao, Qi~Guo, Wenkai He, and Wen Ji.
\newblock Perllm: Personalized inference scheduling with edge-cloud collaboration for diverse llm services.
\newblock {\em arXiv preprint arXiv:2405.14636}, 2024.

\bibitem[\protect\citeauthoryear{Yao \bgroup \em et al.\egroup }{2024}]{yao2024tree}
Shunyu Yao, Dian Yu, Jeffrey Zhao, Izhak Shafran, Tom Griffiths, Yuan Cao, and Karthik Narasimhan.
\newblock Tree of thoughts: Deliberate problem solving with large language models.
\newblock {\em Advances in Neural Information Processing Systems}, 36, 2024.

\bibitem[\protect\citeauthoryear{Ye{\c{s}}il \bgroup \em et al.\egroup }{2024}]{yecsil2024adaptive}
{\c{C}}a{\u{g}}r{\i} Ye{\c{s}}il, Berhan~Turku Ay, Funda~Ay Ak, {\"O}yk{\"u}~Berfin Mercan, and O{\u{g}}uzhan Nefeso{\u{g}}lu.
\newblock Adaptive batch budget for llm inference.
\newblock In {\em 2024 9th International Conference on Computer Science and Engineering (UBMK)}, pages 219--223. IEEE, 2024.

\bibitem[\protect\citeauthoryear{Z \bgroup \em et al.\egroup }{2024}]{zheng2024red}
Zheng Z, Ren X, and et~al Xue~F.
\newblock Response length perception and sequence scheduling: An llm-empowered llm inference pipeline.
\newblock {\em Advances in Neural Information Processing Systems}, 36, 2024.

\bibitem[\protect\citeauthoryear{Zheng \bgroup \em et al.\egroup }{2024}]{zheng2024response}
Zangwei Zheng, Xiaozhe Ren, Fuzhao Xue, Yang Luo, Xin Jiang, and Yang You.
\newblock Response length perception and sequence scheduling: An llm-empowered llm inference pipeline.
\newblock {\em Advances in Neural Information Processing Systems}, 36, 2024.

\bibitem[\protect\citeauthoryear{Zhuang \bgroup \em et al.\egroup }{2024}]{zhuang2024toolqa}
Yuchen Zhuang, Yue Yu, Kuan Wang, Haotian Sun, and Chao Zhang.
\newblock Toolqa: A dataset for llm question answering with external tools.
\newblock {\em Advances in Neural Information Processing Systems}, 36, 2024.

\end{thebibliography}

\end{document}